# Can an Old Dog Be Taught New Tricks? Taking LLMs Beyond Sentence Level Translation

Alaina Brandt 0009-0004-0530-203X

**Abstract.** Automatic translation systems, from CAT tools to MT, overwhelmingly treat translation as a sentence-by-sentence act. This paper asks whether LLMs can be moved beyond that paradigm through whole-document, corpus-informed translation. We present PAT (Pragmatic Auto-Translator), a RAG-based system that pairs user-configured specifications with context from a comparable corpus of authentic longform texts in U.S. English and Latin American Spanish, passing retrieved paragraph-, section-, and document-level examples to an LLM for whole-document generation. The goal is draft translation for professional verification: target texts reformulated to fit their Spanish-language context, where discourse organization, rhetorical style, and pragmatic norms differ meaningfully from English. We evaluated six automatic translations of essays on generative AI across three projects using a customized MQM typology, assessed by two trained evaluators working from U.S. English into LATAM and Mexican Spanish. Results show that a limited prompt produced no meaningful reformulation, and specifications and corpus-informed translations at times showed substantial reformulation, though not always to effect. We find that LLMs can be moved toward reformulation and away from the sentence-by-sentence paradigm, though more work is needed to improve the effectiveness of those reformulations. In this paper, we discuss considerations related to automatic translation system design, corpus construction, and translation quality evaluation methodology and results.

**Keywords:** sentence-by-sentence translation, whole-document translation, corpus-informed translation, pragmatic equivalence, automatic translation, machine translation

## 1. Introduction

Professional translation follows a workflow of source text analysis, the establishment of specifications (work requirements) [1], and specialized research that then informs the production of a translation. Our research asks if automatic whole-document translation performed by LLMs improves within systems that follow a similar workflow. We consider improvement to be the production of natural sounding texts with necessary reformulations at the whole document level that appropriately convey source text meaning, given audience and purpose [2].

To simulate the specialized research a professional translator carries out, we built a RAG-based translation pipeline informed by a comparable corpus, where the text is translated by an LLM on a whole document level to prevent issues related to accuracy and naturalness that stem from sentence-by-sentence translation [3]. We then conduct quality evaluations on the translations based on international standards like MQM [4].

In this paper, we first establish that without guidance, LLMs produce sentence-by-sentence translation (Section 3). We then present the architecture of our corpus-informed translation system, PAT (Section 4), and our evaluation project setup (Section 5), followed by observations from the translation quality evaluations conducted by trained evaluators using a customized MQM typology (Section 6). We conclude with observations about the impact of specifications- and corpus-informed approaches on automatic translation quality and recommendations for further research. Before presenting the research study, Section 2 reviews relevant literature from translation studies, contrastive rhetoric, and automatic translation.

## 2. Background

Within traditional production environments, translation quality is measured as whether a product was produced at the desired cost and within the desired timeline [5], while meeting additional requirements established in specifications [6]. Cost is foregrounded here as translation is most often produced on demand for commodity markets, where the systems and architectures in which it is performed have downward pressure on its value [7, 8, 9].

Word/phrase-level (TBX [10]) and sentence-level (XLIFF, TMX [11, 12]) estimates and controls are the trend within the language industry, where content is split up into word/phrase/sentence-pair level units that are stored in databases to be applied in translation as long as the words match. This practice of segmentation extends to automatic translation, where sentence-to-sentence modeling and quality measurement are pervasive [13, 14, 15]. The problems here are that the statistical word matching underlying the CAT approach bears no relation to semantic equivalence or stylistic credibility, and that sentence-in, sentence-out methods, more broadly, don't account for adequacy at the whole-text level.

Much could be said about how adequacy is understood within translation studies and for the purposes of our research. In the history of translation, adequacy has been seen as fidelity to the form (within the academic discipline of grammar in Antiquity [16], *les belles infidèles* that prioritized French literary form over rendering meaning accurately [17]) to correspondence in the effect (sense-for-sense translation) [18, 19] and visibility instead of erasure of actors and cultures [20] (and so on). Where most technological systems continue in the vein of sentence-by-sentence fidelity to the source's form, translation scholarship asks for more: that a translation effectively conveys the author's ideas, evokes their style, navigates the differences in how each language organizes ideas, addresses conceptual and cultural gaps, and produces an intentional target text for its audience. This points to the desire for pragmatic equivalence in translation.

Contemporary translation scholars and researchers describe pragmatics in complementary ways. In *The Pragmatic Translator* book, pragmatics is defined as language in relation to its context, where three overlapping dimensions are at play: performative (effects), interpersonal (people involved), and locative (place and time) [21]. Baker presents pragmatics as a matter of the coherence of a text with the expectations of its readers: "A network of relations which is valid and makes sense in one society may not be valid in another" (232). Here, we learn that reformulation that works well beyond how words are ordered in a sentence is often appropriate and necessary, where such adaptations can include additions, omissions, restructuring of the argument, and so on [22]. As such, pragmatic translation does the necessary rewriting and reorganization that makes texts fit their new context.

To illustrate the concept of pragmatic equivalence, we'll discuss specific aspects of it that manifested in our research. We tested translation quality of essays translated from English to Spanish in the domain of generative AI. We worked primarily with English to Spanish (and not vice versa) because LLMs are biased toward better performance in English, due to the over-representation of English in the training data [23]. In terms of the discourse patterns made evident through contrastive rhetoric broadly, our project works between a language that tends to be expository and linear in nature (English) and one that can be extraneous with complex digressions (Spanish) [24]. These two languages organize paragraphs, argumentative flow, and text types differently [25], differences that have been shown to hold true even between variants of the same language [26], on top of the obvious difference of vocabulary sets. What is considered an acceptable length differs as well, with Spanish being recognized in the language industry as being more verbose.[1] In terms of the relationship of a text to its context, our work with just three texts presented challenges such as there being no equivalent genre (Sex-In-The-City-style dating columns), that very different concept systems were being activated for the culturally-embedded vocabulary of cuss words, in a text type that prioritizes stylistic beauty.

We conduct our research alongside a growing body of work on document-level automatic translation. Post & Junczys find the continued focus on the sentence-by-sentence paradigm to be unfortunate, particularly when LLMs are equipped to process whole documents. Their paper presents solutions to the impediments preventing the wider adoption of whole-document translation, such as the availability of datasets [3]. Wang et al. found that providing longer document-level input to ChatGPT improved translation quality and discourse awareness through prompting alone, and that while commercial systems such as Google Translate and DeepL outperformed LLMs in BLEU scores, GPT-4 and 3.5 outperformed these systems in human evaluations. They attribute this to the simplistic nature of the BLEU measure, where only n-gram similarity between the output translation and a reference translation is taken into consideration, whereas human evaluators also consider factors like "coherence, fluency, and naturalness" [27]. Building on this, Sun et al. found that a single-pass prompt instructing a model to do document level translation is more effective in producing "fluent, readable, and cohesive output," citing similar problems with BLEU scores favoring literal sentence-level translations [28].

[1] Language industry metrics point to a ~1:1.2-1.3 length difference between English and Spanish.

## 3. Demonstration

Our work assumes that without guidance, LLMs produce sentence-level translation. Comparing translation outputs from several models that have translated the same text twice, once with a limited prompt ("translate this text") and a second time with a reformulation prompt instructing the model not to translate sentence-by-sentence, leads us to believe this to be true. We considered five texts translated twice each by Claude Sonnet 4.6, DeepSeek V4 Flash, Gemini 2.5 Flash, and Gemini 2.5 Pro, producing forty translations in total. When the limited prompt was used, all twenty translations showed what we call here close translation: roughly sentence-level translation that follows source phrasing. When the reformulation prompt was used, models produced freer translations with sentence and paragraph reformulations thirteen out of twenty times, though freeness sometimes came with sizable omissions. Model performance varied, with some showing more receptiveness to reformulation than others; we discuss possible causes for this in *Subsection 4.5*. Results are summarized in *Table 1*.

**Table 1.** Model performance translating five texts from English to Spanish with our system prompt. All twenty translations produced from a limited prompt ("translate this text") showed close translation. Thirteen of twenty produced from the reformulation prompt showed receptiveness to reformulation, at varying degrees. Models tested: Claude Sonnet 4.6, DeepSeek V4 Flash, Gemini 2.5 Flash, and Gemini 2.5 Pro (neither Claude nor DeepSeek in extended thinking mode). Note: only ten of the forty total translations reflected expected English-to-Spanish text expansion (industry standard: ~1:1.2–1.3).

| Model used | Freer translations out of 5 projects | Nature of freer translations |
|---|---|---|
| Claude Sonnet 4.6 (Low) | 4 | Moves the most toward rewriting, makes sizable omissions |
| DeepSeek V4 Flash | 3 | Sometimes starts freer, and then settles back into close translation |
| Gemini 2.5 Flash | 2 | Subtler shifts to more meaningful reformulations |
| Gemini 2.5 Pro | 4 | Subtle through meaningful shifts, with some generalizing and streamlining of the text |

## 4. System architecture

This section describes the architecture of the PAT system, through which we produce corpus-informed translation. It's important to note that even if the LLM generates an appropriately reformulated translation, the PAT system produces what ASTM F2575 refers to as unverified translation (rather than professionally verified translation) [1].

### 4.1 Corpus

Corpora are foundational to the translation approach described here. In professional environments, translators use corpora to maintain subject-field expertise, substantiate linguistic choices (vocabulary, collocations, elements of style) [29] and identify appropriate specialized terminology [30]. Our hypothesis is that LLM generation can be guided similarly: that sharing writing samples illustrating pragmatic equivalence in the prompt will push the model toward more appropriate outputs than it would produce otherwise. To that end, the corpus we built for this research contains only "authentic" writing (no translations, no synthetic text) to avoid introducing the source-language interference patterns known to appear in translated text [31] and to counteract the bias toward English that stems from LLMs' training data [32]. In practice, some translated quotes and AI-generated chat excerpts have made their way in through the texts themselves, which we note as a limitation.

For the purposes of this research, we built our PAT-GAI-Longform-ESP-419-ENG-USA corpus. Our initial intention was to build a Mexican Spanish <> U.S. English comparable corpus, but English writing on the topic outpaces Spanish writing. We therefore expanded our corpus to prioritize texts from LATAM. We also include select content in Castilian Spanish and international English. The corpus contains essays, reports, journal articles, conference papers, magazine articles, reviews, book chapters, legal commentary, and so on. At the time of writing the corpus contains ~200,000 words in English and ~225,000 words in Spanish.

In keeping with calls for transparency and standardization, the characteristics and uses of this dataset are described in a datasheet whose structure is based upon the template proposed by Gebru et al. [33]. Information on how to access that datasheet can be found in *Annex 1*.

### 4.2 Embeddings

Retrieval in our context needs to be semantically, pragmatically and discourse sensitive. Querying the embeddings should derive comparable multilingual texts in the source and target languages that are similar to the source text in terms of the ideas conveyed and the writing style.[2] Representing texts at multiple levels of document structure has been shown to capture richer semantic information than single-level approaches in the context of trained matching models [34]. PAT extends this principle to pre-trained embeddings by indexing corpus texts at paragraph, section, and document levels, so that retrieved texts are segmented to serve as examples of features like terminology usage in context and sentence-level organization (paragraph level), cohesion and discourse structure (section level), and overall text organization (document level). A document-level embedding of the source text is then compared against the corpus embeddings during retrieval. The user can select from among the examples retrieved those that will be passed to the translation model.

Among the models available for embedding long-form texts onto a multilingual vector space and performing cross-lingual retrieval, we selected `jina-embeddings-v3`. At 570M parameters, it's a lightweight, lower latency model with competitive performance relative to larger models. It offers a context length of ~8k tokens and output dimensions of 1024, which meet (or, in the case of paragraph-level embeddings, exceed) PAT needs. It also scored well on Multilingual MTEB (average: 64.44) and LongEmbed (average: 70.39) tasks [35], and Jina offers a generous free API allocation for research use.

The model also has shortcomings worth noting. Its base model Facebook AI's `xlm-roberta-large` [36], was pretrained with sentence-level masked language modeling [37], meaning it has no pre-training signal for how sentences flow within paragraphs or how paragraphs flow within a text. The field has also advanced since our 2025 selection: similar models that are now available process ~32k tokens at comparable parameter counts (~596M) and the same output dimensions (1024). More recent comparable models outperform jina-v3 across task groups on the Multilingual MTEB (v2) [38], with `jina-embeddings-v5-text-small` being a straightforward upgrade and `microsoft/harrier-oss-v1-0.6b` being a standout alternative (see *Annex 2* for a fuller comparison). `jina-v5` was also trained on the `Qwen3-0.6B-Base` model [39], a compelling fit for our use case, since its pre-training included a dedicated long-context stage on documents of 4,096-32,768 tokens, giving the base model exposure to intra-document coherence patterns absent from sentence-level pre-training [40].

[2] We treat it as a design assumption that document-level embeddings encode stylistic alongside propositional similarity. PAT's system design, in which retrieved embeddings are saved, positions us to test this assumption directly.

Of the 131 tasks against which embedding models are measured on the MMTEB, none appears to address long-form to long-form retrieval directly. This would be the use case most relevant to PAT, where whole documents and extended passages are compared rather than short queries against passages. This points to a gap in current benchmarking for retrieval.

### 4.3 System Prompt

Research shows that specifications can improve LLM translation quality [41, 42]. Within PAT, user-selected specifications and corpus context are brought together in a system prompt that instructs the LLM how to use the resources provided during translation: the specifications tell it how and for whom to reformulate, and corpus context illustrates what that can look like in practice.

*Specifications*
When a source text is submitted to the PAT system, users can configure their request by setting specifications. These can also be auto-configured through a call to an LLM. This means that PAT is compliant with the ASTM F2575 standard (Standard Practice for Language Translation) in that the system is designed to include specifications with every translation request [1]. Of possible translation parameters [43, 44], users have the opportunity to delimit the subject field, source and target language variants, the text type, its purpose, the author's perspective (from what point of view do they write), the audience, and distribution scope. Users can also include content complexity notes on translation challenges they have identified through their own critical reading of the source text for cases where they do not want to leave specific features to the probabilistic outputs of the models.

*Prompt engineering*
The word "translate" appears to trigger sentence-level processing in LLMs, reflecting how pervasively sentence-by-sentence translation is represented in their training data (cf. *Section 3*). Our prompt engineering decisions were shaped in part by this constraint. First, we tested sidestepping the word "translation" to avoid triggering associations with sentence-level translation (using words like "rewriting" instead). We found that repeatedly instructing the model not to perform sentence-by-sentence translation was more effective, at least for the Deepseek V4 and Gemini 2.5 models we tested. Second, the prompt instructs the model to read and comprehend the full source text before generating output, in an attempt to counteract the tendency toward sentence-by-sentence translation that emerges when models proceed directly to generation.

Information on how to access our system prompt can be found in *Annex 1*.

### 4.4 Large language model

Translation is an emergent capability in large language models rather than an explicitly trained one, arising from exposure to vast quantities of whole texts. This creates more room to do translation differently than it has been done in translation technology: unlike NMT models trained on parallel corpora for sentence-by-sentence output, LLMs are not structurally predisposed to that paradigm.

Gemini 2.5 Pro was selected as PAT's translation engine on the basis of its receptiveness to whole-document reformulation (cf. *Table 1*). To illustrate how training methodology may influence receptiveness, we contrast it with DeepSeek V4 Flash. Both are long-context, mixture-of-experts models with 1M token context windows, but they optimize for different objectives. DeepSeek V4 forgoes RLHF in favor of automated rubric-guided reinforcement learning designed to reward verifiable correctness in domains like math, code, and logic [45]. Gemini 2.5 Pro incorporates human preference data alongside a critic model and human feedback in a methodology referred to as Reinforcement Learning from Human and Critic Feedback (RL*F) [46]. The result is a model that prioritizes helpfulness and instruction following. These qualities matter when the task requires sustained reformulation rather than convergence on a ground truth answer. The pattern in *Table 1* where DeepSeek sometimes begins with reformulation before reverting to close translation is consistent with a model that allocates compute efficiently toward correctness rather than maintaining an open-ended rewriting stance across a long generation.

In our work, the Gemini 2.5 Pro model was run at default temperature (1.0) and top-P (0.95) values [47] so that reformulations in the output could be attributed to system design rather than to widened sampling probabilities.

## 5. Evaluation Design

To prepare for and conduct evaluations from which we can make conjectures about how specifications- and corpus-informed approaches affect LLMs' translation quality, we built eight projects for evaluation on Label Studio, evaluated by trained evaluators using a customized MQM typology.

### 5.1 Quality Evaluators

We conducted translation quality evaluations with a group of seven evaluators working between English and Spanish, heterogeneous in translation direction and Spanish language variant as a group. All evaluators completed the "Loc401: TQMS Based on MQM" course alongside participation, and calibration sessions were held between projects to work toward harmonization.[3] Four evaluators withdrew unexpectedly, and a fifth completed over half the projects but showed substantially lower agreement and so was not retained for the remainder of the research study; their annotations are excluded from our quality analysis.

The quality analysis presented in the next section is based on the work of two evaluators who work from U.S. English to Mexican Spanish. These evaluators share the same educational background and prior exposure to MQM.

### 5.2 Evaluation projects

All eight TQE projects were between Spanish<>English in the domain of generative AI. We selected or adapted texts for translation of 1200 words. Projects 0-4 were part of our training phase, in which the goal was for evaluators to gain experience applying our MQM typology. Projects 5-7 were part of our production phase, from which we make observations about the quality of automatic translations produced under different conditions. For the production phase projects, evaluations were done on the automatic translation of essays produced by PAT. TQE-Project-5 into LATAM Spanish tested specifications-informed translation (from our system prompt) against translation produced from a limited prompt.[4] TQE-Projects-6 and 7 into Mexican Spanish tested corpus-informed translation. All four texts from the corpus-informed translation tests were produced from our reformulation (i.e. system) prompt.

All evaluation projects were completed on a private Label Studio instance [48], which allows annotation of whole, unsegmented texts using a customized labeling setup. Evaluators were instructed not to view one another's annotations until after submitting their own work.

### 5.3 Customized MQM typology/holistic quality measures

Our labeling setup is based on two standards: MQM Core for individual error annotation [49], and ASTM WK54884, a draft standard on holistic translation quality evaluation [50]. Where WK54884 uses the term "adequacy," our setup retains the earlier term "correspondence."

[3] https://loc401.locessentials.com/#content/eng-usa/introduction.md
[4] The LATAM Spanish target in this project was selected to accommodate an evaluator specializing in Argentine Spanish who subsequently withdrew. Had we anticipated this, all production phase projects would have targeted Mexican Spanish. That is to say, the variant difference here is incidental rather than intentional.

For individual error labeling, we developed a customized typology based on MQM Core. For each error marked, evaluators were to select the error category and specific error type, assign an impact level (ranging from neutral to showstopper), and leave a comment explaining the error. The most significant customization was the addition of Cohesion as an error type under Accuracy, to capture errors related to missing or insufficient signposting that guides a reader through the text. For later projects, we introduced a Decision Tree to help evaluators navigate category and error type selection, based on materials developed to complement MQM [51].

As a holistic quality measure, evaluators rated the target text's overall correspondence to the source and its readability as a standalone target, each on a 4-point scale, with space to comment on these two dimensions along with document-level issues.

### 5.4 Agreement Calculations

Interannotator agreement is an important indicator of evaluation validity. In translation, however, just as any source text can have many correct translations, any translation can have many correct evaluations. This reality sets a natural ceiling on agreement. We therefore set a reliability target of 60% rather than the higher thresholds common in other annotation tasks. Partial span agreement was calculated as an F1 score at the character level [52]. For each annotator pair, every character position covered by at least one annotator's error spans was treated as a binary indicator of whether that character had been flagged. Precision was the proportion of character positions labeled by Annotator A that were also labeled by Annotator B; recall was the reverse. The F1 score is the harmonic mean of the two.[5]

## 6. Interpretation of Results

### 6.1 Reformulation

Before turning to the translation quality evaluations, we assess whether the production-phase translations show evidence of reformulation at all. Our findings are that a limited prompt produced no meaningful reformulation; specifications alone produced minor to substantial reformulation; and specifications combined with corpus context produced moderate to substantial reformulation. Notably, the specifications-only translation targeting the broader LATAM Spanish variant showed less reformulation than those targeting Mexican Spanish specifically, which may reflect the greater neutrality required when addressing a wider regional audience rather than a more specific one. While corpus-informed translations showed the strongest reformulation overall, the specifications-only condition in Projects 6 and 7 also produced moderate to substantial reformulation, suggesting that the system prompt may be an important driver of reformulation. *Table 2* summarizes our observations across the six texts.

**Table 2.** Degree of reformulation observed across six production-phase automatic translations, assessed independently of translation quality evaluations. "Prompt" indicates specifications-informed translation produced from the PAT system prompt; "No prompt" indicates translation produced from a limited prompt; "Context" indicates specifications- and corpus-informed translation. **Reformulation key:** None / Minor / Moderate / Substantial

| Project & Text | Reformulation | Observations |
|---|---|---|
| TQE-5 Text 1: eng-USA → esp-LATAM - Prompt | Minor | Two short paragraphs combined into one, some sentences combined to form longer ones; some addition of signposts, phrasing reformulations |
| TQE-5 Text 2: eng-USA → esp-LATAM - No prompt | None | Translation mostly even follows source structure at the phrase level, treating comma-delimited units as discrete translation segments |

[5] We also calculated error category agreement restricted to partial span matches. Scores remained below 50% agreement across all projects, falling short of our reliability target. This is touched upon briefly in *Section 6*.

| TQE-6 Text 1: eng-USA → esp-MEX - Context | Substantial | Introductory statement expanded into a full paragraph; short sentences elaborated; a long paragraph containing two discrete ideas divided into two; ideas reordered within paragraphs; some source repetition omitted |
|---|---|---|
| TQE-6 Text 2: eng-USA → esp-MEX - Prompt | Moderate | Reformulations comparable to TQE-6 Text 1 in most types but not degree; idea reordering absent |
| TQE-7 Text 1: eng-USA → esp-MEX - Prompt | Substantial | Short paragraphs consolidated; some reorganization of ideas; instances of creative invention |
| TQE-7 Text 2: eng-USA → esp-MEX - Context | Moderate | Extensive paragraph consolidation; less reorganization of ideas than TQE-7 Text 1 |

**6.2 Corpus Context**

The corpus-informed translations show evidence that retrieved context shaped both the style and the degree of reformulation in the output, though the strength of that influence varied and appears sensitive to how context was delivered.

In Project 6, nine context passages were selected from retrieval results: three paragraphs each in English and Spanish, and three section-level passages in Spanish. The English paragraphs illustrated relatively shorter, more direct sentence structures typical of the language; the Spanish passages demonstrated the more complex, digression-tolerant syntax characteristic of Spanish essayistic writing, including frequent use of signposting expressions (*por lo tanto, en cambio, en la actualidad, en primer lugar*) and comma-offset subordinate clauses. Both translations showed evidence of these same features, but the corpus-informed translation did to a greater degree and with more creative latitude. For example, where the prompt-only translation rendered “bloggers” as *blogueros* and “devoted a lot of energy” as *han dedicado mucho tiempo y energía*, the corpus-informed translation rendered the same phrases as *un sinfín de blogueros* and *han dedicado ríos de tinta*, idiomatic formulations consistent with the register and rhetorical style of the Spanish corpus texts. The corpus-informed translation also showed more structural reorganization: ideas reordered within sentences, source repetition omitted, and sentence boundaries shifted to produce longer, more integrated units. The prompt-only translation was reformulated as instructed but stayed closer to source phrasing throughout. Selected comparisons are shown in *Table 3*.

Project 7 also presents an interesting picture. Eight context passages were selected: one English section illustrating Substack-style personal essay writing (chosen because the source text’s Sex-in-the-City-style dating column genre has no ready equivalent in Spanish) and seven Spanish passages offering related thematic perspectives on the source topic of a woman discovering her boyfriend’s complaints about her to ChatGPT. Spanish contexts included an observer’s account of a friend using ChatGPT as a relationship counselor, song lyrics about staying with a partner ChatGPT recommended leaving, a student’s account of using AI for therapy, and so on. Here, however, the corpus-informed translation stayed closer to the source text’s organizational structure than the prompt-only translation did, consolidating the source’s 45 paragraphs into 35, compared to 42 in the prompt-only version. This pattern of greater paragraph consolidation but less structural reorganization suggests that the volume of long-form context passed to the model may have diffused attention away from reformulation rather than reinforcing it. Context was passed as concatenated blocks of text with paragraph breaks invisible to the model, a formatting issue since resolved in the system. This is noted as a limitation of the current evaluation.

Taken together, the Project 6 and 7 results suggest that corpus context can meaningfully influence translation style and enable more target-language-idiomatic reformulation, but that retrieval quantity and context formatting matter. Carefully selected, well-formatted context appears to give the model something to imitate; excessive or poorly delimited context may instead function as noise.

**Table 3.** Comparisons of corpus-informed and prompt-only English-to-Mexican-Spanish translations of [53] in project 6, illustrating differences in reformulation degree and target-language idiomaticity.

| ST | TT - Corpus informed | TT - Not corpus informed |
|---|---|---|
| Typically, we talk about algorithms that are implemented by a computer and written in computer code. But algorithms can also be written in math, like the quadratic formula or the equation to calculate the area of a circle; or they can be written in natural language, like a chocolate chip cookie recipe or instructions for assembling a desk. | Los vemos en el código con el que se programa una computadora**, sí,** pero también en el lenguaje de las matemáticas **—como la fórmula cuadrática o la ecuación para calcular el área de un círculo—** e incluso en el lenguaje cotidiano, como los pasos de una receta para hacer galletas o el instructivo para armar un mueble. | Generalmente pensamos en algoritmos implementados por una computadora y escritos en código**, pero en realidad están por todas partes.** Pueden ser una fórmula matemática, como la que usamos para resolver una ecuación de segundo grado o para calcular el área de un círculo; o pueden estar escritos en nuestro propio idioma, como la receta para hacer galletas con chispas de chocolate o las instrucciones para armar un mueble. |
| Like the dog, machine learning is unable to provide reasons and explanation for how it does things. | El aprendizaje automático**, como el perro,** es incapaz de ofrecer razones o justificaciones sobre cómo hace lo que hace. | Al igual que el perro, el aprendizaje automático es incapaz de darnos las razones o explicaciones de cómo hace las cosas. |
| Science fiction authors have written about the singularity for a long time. Philosophers, ethicists, technologists, and people with blogs have devoted a lot of energy and time to fearing or not-fearing the singularity. | Autores de ciencia ficción, filósofos, tecnólogos **y un sinfín de blogueros** han dedicado **ríos de tinta** a temer **—o no—** este escenario. | Autores de ciencia ficción, filósofos, tecnólogos y blogueros han dedicado mucho tiempo y energía a debatir sobre este concepto. |

### 6.3 Interannotator Reliability

*Table 4* presents span match agreement scores between the two quality evaluators across all production-phase projects. Scores ranged from 49.5% to 61.4%. The line graph (see Fig. 1) shows agreement across seven TQE-2026-1 projects over time, including the training phase. Category agreement did not approach our 60% reliability target, an expectable outcome, as moving from span agreement to category agreement requires a higher level of annotation precision.

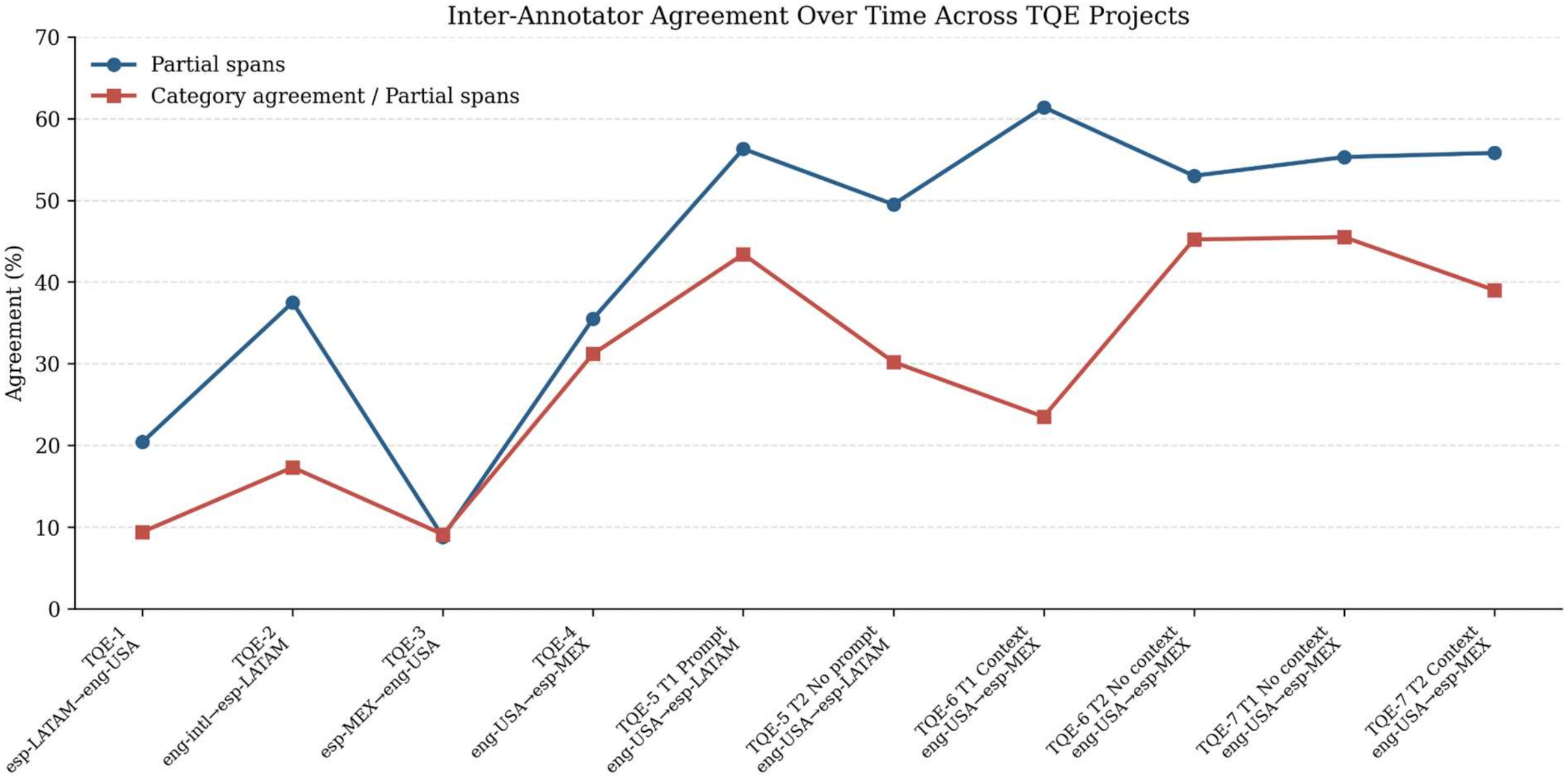

**Fig. 1.** Partial span agreement (F1, character-level) and category agreement as a proportion of partial span matches, measured between two evaluators across seven TQE projects in chronological order. Training-phase projects (TQE-1 through TQE-4) include into-English and into-Spanish directions; production-phase projects (TQE-5 through TQE-7) are exclusively into-Spanish. The upward trend in partial span agreement across the production phase reflects improved evaluator alignment over time.

Several observations can be made about span agreement. The into-English training projects produced the lowest agreement scores for evaluators whose A language is Spanish, consistent with the language-industry expectation that translators work most reliably into their dominant language. Setting those projects aside, partial span agreement across the four into-Spanish production projects ranged from 49.5% to 61.4%, an improvement over the ~35% scores seen in the two into-Spanish training projects. This may reflect the reduced evaluator pool: working as a pair rather than within a larger group, the two evaluators no longer needed to accommodate a wider range of interpretive perspectives. Calibration sessions across projects may have contributed as well.

The raw span counts add relevant context. Labeled spans across the production-phase texts were 83 and 86 (Prompt/No Prompt) for Project 5, 51 and 31 (Context/Prompt) for Project 6, and 44 and 41 (Prompt/Context) for Project 7. The volume of errors flagged is itself an argument for multiple evaluators on error-heavy texts, since divergence in annotation is more likely when there is more to find and interpret. We are also reminded that translation quality evaluation has no ground-truth answer: measuring against a gold standard would only be reliably predictive for errors requiring no interpretation, such as number, punctuation, or spelling errors. Judgment calls, by their nature, resist standardization.

The following subsection addresses the question of whether specifications- or corpus-informed conditions produced better automatic translations according to the evaluations, with the caveat that findings rest on moderate partial span agreement between two evaluators.

**Table 4.** Span match agreement scores (F1, character-level) between two evaluators across six production-phase automatic translations. "Prompt" denotes specifications-informed translation produced from the PAT system prompt; "No prompt" denotes translation produced from a limited prompt ("translate this text"); "Context" denotes corpus-informed translation produced from the PAT system prompt. All corpus-informed translations were also specifications-informed.

| Project | Span agreement |
|---|---|
| TQE-5 Text 1: eng-USA → esp-LATAM - Prompt | 56.3% |
| TQE-5 Text 2: eng-USA → esp-LATAM - No prompt | 49.5% |
| TQE-6 Text 1: eng-USA → esp-MEX - Context | 61.4% |
| TQE-6 Text 2: eng-USA → esp-MEX - Prompt | 53% |
| TQE-7 Text 1: eng-USA → esp-MEX - Prompt | 55.3% |
| TQE-7 Text 2: eng-USA → esp-MEX - Context | 55.8% |

### 6.4 Quality Evaluations

*Table 5* presents total labeled spans across both evaluators alongside mean correspondence and readability scores for each production-phase translation. To identify what the evaluations point to as the better translation within each project pair, we look for the lower span total and the higher correspondence and readability scores. By these measures, evaluators rated the specifications-informed translation as better in TQE-5, the prompt-only translation as better in TQE-6, and the corpus-informed translation as better in TQE-7. Correspondence and readability scores did not contradict one another across any project, and error counts align broadly with evaluator preferences: translations that received lower correspondence and readability scores also accumulated more labeled spans.

The more striking pattern is that in both corpus-informed projects, evaluators preferred the translation with less reformulation. In TQE-6, the prompt-only translation was rated higher despite showing less structural reorganization and closer source adherence than the corpus-informed version. In TQE-7, the corpus-informed translation was rated slightly higher, but it was also the least reformulated of the two, consolidating paragraphs without substantially reorganizing ideas. From this we conjecture that evaluators trained in an industry with a strong bias toward sentence-by-sentence translation may carry a corresponding bias in their quality assessments, even when freer reformulation would better serve the target audience. This would be worth testing systematically with a larger evaluator pool and better calibration on reformulation as a positive quality signal.

**Table 5.** Total labeled spans (summed across two evaluators) and mean correspondence and readability scores for six production-phase automatic translations. Correspondence is rated on a 1-4 scale (1 = major meaning differences; 4 = excellent correspondence). Readability is rated on a 1-4 scale (1 = difficult to read; 4 = reads naturally). "Prompt" denotes specifications-informed translation; "No prompt" denotes limited-prompt translation; "Context" denotes corpus-informed translation produced from the PAT system prompt.

| Translation | Total Spans | Mean correspondence | Mean readability |
|---|---|---|---|
| TQE-5 Text 1: eng-USA → esp-LATAM - Prompt | 83 | 3 | 2 |
| TQE-5 Text 2: eng-USA → esp-LATAM - No prompt | 86 | 2 | 1 |
| TQE-6 Text 1: eng-USA → esp-MEX - Context | 51 | 2.5 | 2.5 |
| TQE-6 Text 2: eng-USA → esp-MEX - Prompt | 31 | 3 | 3 |
| TQE-7 Text 1: eng-USA → esp-MEX - Prompt | 44 | 2.5 | 2.5 |
| TQE-7 Text 2: eng-USA → esp-MEX - Context | 41 | 2.5 | 3.0 |

*TQE Project 5 - Discussion*

Both evaluators rated the limited-prompt translation's readability as 1 out of 4 and its correspondence as 2 out of 4. The specifications-informed translation fared better on both measures, with correspondence rated 3 out of 4 by both evaluators, though one evaluator rated readability as 1 out of 4, noting that while the text was functional, it fell short of the naturalness required for a persuasive argumentative text. The limited-prompt translation was characterized as terrible to read.

Despite the overall criticism of both translations, the specifications-informed translation is the stronger of the two, and its use of specifications aligns with broader industry standards requiring them within translation workflows, particularly ASTM F2575 [1]. We note that while the "prompt" approach produced a better draft translation, relative to the "no prompt" translation, it still didn't take the reformulation far enough for one evaluator. Notable remaining issues in the system-prompt translation include a small amount of gender agreement errors where, in one instance, the referent appeared in the previous sentence rather than the current one. This error type is characteristic of sentence-by-sentence processing. The model failed to carry information across a sentence boundary. This suggests that even when following a reformulation prompt, the sentence-level bias in LLMs remains difficult to fully overcome.

*TQE Project 6 - Discussion*

Project 6 does not lend itself to straightforward conclusions. The corpus-informed translation shows clear evidence of corpus influence and received lower scores. Evaluator comments point to a specific tension: the corpus pushed the model toward features of Spanish essayistic writing, but without the judgment to apply them effectively. The addition of connectors came along with the loss of important elements of the source text. The reformulations in the prompt-only translation were found to be similarly ineffective, though to a lesser degree, and overall, the prompt-only translation stayed closer to the source and tracked its ideas more faithfully according to evaluator comments.

Select errors that the evaluators flagged are present in *Table 6*. The context-informed translation had 20 more spans marked than the specifications-informed translation. Many of these spans were marked by one evaluator but not the other. Where there was more agreement was in issues present in both texts: additions, literalness, and an omission. The errors marked by only one evaluator in the corpus-informed translation had to do with literalness, flow, register, and grammar usage. The differences in the number of spans marked for each text could have several causes: the volume of errors may have caused evaluators to miss some that their peer caught, the reformulations in the corpus-informed translation may have simply given evaluators more to react to individually without that reaction converging into agreement, or some combination of both.

**Table 6.** Selected errors from TQE Project 6 illustrating patterns across the corpus-informed and prompt-only translations. Entries were selected for interest.

| Span | Notes |
|---|---|
| Addition of introductory paragraph | Marked by both evaluators in both translations. Flagged as translator/model overreach. |
| si… "Juárez" se refiere… a la alcaldía en la capital del país | Both evaluators, corpus-informed translation. The model attempted to localize a reference to Washington, as instructed. The way Juárez was framed in the corpus-informed translation was ambiguous (capital of what country?) and confusing (the alcaldía is Benito Juárez, the colonia is Juárez). Interestingly, in the translation produced from just the system prompt, the framing for the localization to Juárez was not marked as an error. |
| caja negra | Both evaluators, both translations. Flagged as a culturally opaque term with no Spanish Wikipedia entry. |
| una diferencia fundamental: a diferencia de… | Both evaluators, both translations. "Diferencia" appears twice in immediate proximity. Flagged as awkward. |
| Un error aún más profundo… | Both evaluators, corpus-informed translation. Flagged as an interpretation that introduces a negative connotation not present in the source. |
| ríos de tinta | Both evaluators, corpus-informed translation. Both evaluators pointed out that the metaphor of writers dedicating a lot of ink to the singularity does not extend to bloggers, who do not use ink to write. |
| escucha y conversa… | One evaluator, both files. Flagged as pleonastic, since conversations already imply listening (ST: listen to and talk to) |
| Variation in person | One evaluator, corpus-informed translation. The translation shifts back and forth between "you" and "we". Flagged as an inconsistency. |

*TQE Project 7 - Discussion*

Project 7 surfaces a rich set of errors across both translations, and the evaluation data reveals distinctive patterns in each. The prompt-only translation's characteristic failure is addition: the model invented content not present in the source, and evaluators agreed on this pattern across multiple instances. The corpus-informed translation shows the opposite tendency: staying closer to the source but in ways that produced misreading and mishandling, including a culturally loaded generalization and a tone-inverting cuss word choice that erases the source's sarcasm entirely.

What both translations share is a pervasive pattern of literalism that points to the limits of automatic translation for this text type. The source text is a personal essay in a genre with no direct Spanish equivalent, dense with colloquialisms, wordplay, and culturally embedded references. Neither translation resolved these challenges. Some failures reflect vocabulary gaps between the two languages; others reflect English syntactic patterns surfacing in the Spanish output. These findings suggest that for text types demanding stylistic precision and cultural embeddedness, the gap between automatic translation and professional translation remains substantial at present, regardless of the prompt used or the corpus context provided.

**Table 7.** Selected errors from TQE Project 7 across prompt-only and corpus-informed English-to-Mexican-Spanish translations of a personal essay in a Sex-in-the-City-style dating column genre [54]. Errors unique to each translation reflect contrasting failure modes; shared errors across both translations point to the limits of automatic translation for this text type.

| Text | Notes |
|---|---|
| TQE-7 Text 1: eng-USA → esp-MEX - Prompt | The distinctive pattern is the presence of additions, most of which evaluators agreed upon (Solo, Sin pensarlo mucho, paralizada, a pesar de todo, Pero pónganse en mi lugar). Additional notable errors: "almost poetically" was translated as "chiste del destino" where "burla de destino" should have been used for the idea of a "trick of fate"; a major mistranslation occurs: the subject is rendered as resting upon his own body, rather than resting on hers. |
| TQE-7 Text 2: eng-USA → esp-MEX - Context | Errors skew towards misunderstanding or mishandling the source. The translation demonstrates bias in one instance, where a reference to the idea of 'Florence Nightingale taking care of cats' is generalized to *loca*, calling to mind the English stereotype of the crazy cat lady; the prompt-only translation generalized this reference to 'a nurse' caring for the cats. The text also translates the cuss word "No shit", whose implicature is a sarcastic 'you don't say,' with an overly strong Spanish cuss word that doesn't carry over the sarcasm. |
| Both texts | Both tests display similar issues that skew toward literalness. The translation of the colloquial "Like excuse me sir" is either nonsensical for the context or completely garbled. There are many instances of the literal vocabulary choice being the wrong one, such as: *luz cálida* rather than *ojos benévolos*, *brújula moral* where *compás moral* should have been used (variant appropriateness), *torriente de conciencia* for *flujo/corriente de conciencia*, or the intonation of "clinical" meaning with surgical precision in a non-medical context not carrying over to the Spanish *clínico*. Some grammar constructions also demonstrate a bias toward English: the use of explicit *Yo* mirrors the English requirement of an overt subject, whereas Spanish often stylistically prefers conveying the subject through verb inflection alone; the English progressive 'to be doing' construction |

surfaces in the Spanish as *estar besándote*, where the simple infinitive *besarte* would be more natural. The target text cannot replicate the double meaning of "A few lines later came the body of it," where body refers to both the text and her body.

## 7. Conclusion

PAT is designed as a human-centered system: its goal is not to make professional translators obsolete but to automate the research and drafting phases of translation in a way that is professionally driven. Corpus construction, specification configuration, and the selection of corpus examples from retrieval are all user-determined, meaning that translation quality is shaped at multiple points by human judgment before the model generates a single word. This paper establishes the foundational architecture of the PAT system and a translation quality management framework for evaluating its outputs. The finer-grained questions, such as how retrieval performance varies across different combinations of specifications, how specifications-informed and corpus-informed conditions interact at scale, and what fine-tuning on tagged long-form datasets would contribute are the work of studies that can build upon the foundation established here.

The biggest question this research leaves open is the scale of the bias toward sentence-by-sentence translation in LLMs, the language industry, and the machine translation community. It would seem too large to overcome entirely, yet the evaluations presented here show that the influence of specifications and corpus context is demonstrable in the output. We have established that LLMs can be moved toward reformulation. The work ahead is making that reformulation effective.

**Acknowledgements.** This project would not have been possible without the contributions of several colleagues. On the frontend, Mohit Agarwal built the PAT system from our design notes with remarkable intuition and responsiveness, solving technical challenges that shaped the system at every stage. On the backend, our sincere appreciation goes to Ilse Barragán Macías, Pamela Cisneros, and Renata Sofía Pérez Casado, whose commitment, thoroughness, and attention to detail in completing the translation quality evaluations made the analysis presented here possible.

## Annexes

**Annex 1.** This paper makes reference to various external resources related to this research:

- **Pragmatic Auto-Translator (PAT)**: Where we manage our corpora and conduct specifications- and corpus-informed automatic translation. Available at: https://auto-translator.locessentials.com
- **PAT-GAI-Longform-ESP-419-ENG-USA**: PDFs available on our PAT platform with a login (same link as above). OCR-processed and Markdown versions available after contributing to the corpus.
- **TQE-2026-1 project series**: The dashboard with datasets and analysis of the translation quality evaluation projects we ran, along with related guidelines. Available at: https://zenodo.org/records/20954845
    - Guidelines include the system prompt and the corpus datasheet.
    - Datasets and analysis are provided for projects 1-7.
    - Also available at: https://locessentials.github.io/tqe-2026-1/

**Annex 2.** `jina-embedding-v3` scores on task group items of the Multilingual MTEB Multilingual (v2) compared with similar, but more contemporary models on the Leaderboard. `jina-v3` has a token window of ~8k, while contemporary models share the same approximate token window (~32k). All models compared share the same approximate parameters (~596M) and dimension output (1024). Task groups presented here include those with the greatest range in performance among models. **Task group key:** Mean (Task) = average score across all MMTEB tasks; Retrieval = document retrieval from a query; MultiLablClass = multilabel text classification; Class = single-label text classification; Cluster = semantic clustering of texts; Bitext Mining = identification of parallel sentences;

InstructionReranking = reranking of results given an explicit instruction. Instruction Reranking scores are not directly comparable to other task group scores due to differences in metric reporting on the MMTEB leaderboard. **Relative performance indicators:** lolo = very low, lo = low, mid = mid, hi = high, hihi = very high, within the range of models compared.

| Model | Mean (Task) | Retrieval | MultiLabl Class | Class | Cluster | Bitext mining | Instruction Reranking |
|---|---|---|---|---|---|---|---|
| jinaai/jina-embeddings-v3 | 58.37 (lolo) | 55.76 (lolo) | 18.38 (lolo) | 58.77 (lolo) | 45.65 (lolo) | 65.25 (lolo) | -1.34 (lolo) |
| jinaai/jina-embeddings-v5-text-small | 67 (hi) | 64.88 (hi) | 41.97 (hihi) | 71.32 (hi) | 53.41 (mid) | 69.71 (lo) | 1.35 (mid) |
| microsoft/harrier-oss-v1-0.6b | 69.01 (hihi) | 70.75 (hihi) | 26.37 (hi) | 73.88 (hihi) | 54 (hi) | 82.85 (hihi) | 0.81 (lo) |
| Qwen/Qwen3-Embedding-0.6B | 64.34 (mid) | 64.65 (mid) | 24.59 (lo) | 66.83 (mid) | 52.33 (lo) | 72.23 (hi) | 5.09 (hihi) |
| codefuse-ai/F2LLM-v2-0.6B | 62.74 (lo) | 59.3 (lo) | 25.23 (mid) | 64.06 (lo) | 56.58 (hihi) | 70.31 (mid) | 1.39 (hi) |